\documentclass[runningheads]{llncs}
\usepackage{graphicx}

\usepackage{tikz}
\usepackage{comment}
\usepackage{amsmath,amssymb} 
\usepackage{color}
\usepackage{multirow}
\usepackage{subfigure}

\usepackage[misc]{ifsym}

\usepackage[accsupp]{axessibility}  

\begin{document}
\pagestyle{headings}
\mainmatter
\def\ECCVSubNumber{4016}  

\title{Content Adaptive Latents and Decoder for Neural Image Compression}

\titlerunning{Content Adaptive Latents and Decoder for Neural Image Compression}
%
\author{Guanbo Pan\inst{1} \and 
Guo Lu\inst{2} \and
Zhihao Hu\inst{1} \and
Dong Xu \inst{3}\textsuperscript{(\Letter)}}
%
\authorrunning{G. Pan et al.}
%
\institute{School of Software, Beihang University, Beijing, China \\
\and
School of Computer Science and Technology, Beijing Institute of Technology, Beijing, China \\
\and
Department of Computer Science, The University of Hong Kong, Hong Kong, China \\
\email{dongxu@cs.hku.hk}
}
\maketitle

\begin{abstract}
In recent years, neural image compression (NIC) algorithms have shown powerful coding performance. However, most of them are not adaptive to the image content. 
Although several content adaptive methods have been proposed by updating the encoder-side components, the adaptability of both latents and the decoder is not well exploited. 
In this work, we propose a new NIC framework that improves the content adaptability on both latents and the decoder. 
Specifically, to remove redundancy in the latents, our content adaptive channel dropping (CACD) method automatically selects the optimal quality levels for the latents spatially and drops the redundant channels. 
Additionally, we propose the content adaptive feature transformation (CAFT) method to improve decoder-side content adaptability by extracting the characteristic information of the image content, which is then used to transform the features in the decoder side. 
Experimental results demonstrate that our proposed methods with the encoder-side updating algorithm achieve the state-of-the-art performance. 
\keywords{Neural Image Compression, Content Adaptive Coding}
\end{abstract}

\section{Introduction}
Data compression has been studied for decades as an essential issue to alleviate data storage and transmission burden. The traditional codecs, such as JPEG \cite{jpeg}, JPEG2000 \cite{jpeg2000}, BPG \cite{bpg} for image compression and H.264 \cite{h264}, H.265 \cite{h265}, H.266 \cite{h266} for video compression, still prevail nowadays. In recent years, neural image compression (NIC) has shown promising coding performance due to its powerful nonlinear transformation capability and end-to-end optimization strategy. The recent state-of-the-art NIC methods like \cite{acm21} outperform the latest traditional compression standard Versatile Video Coding (VVC) \cite{h266} on various datasets including the Kodak \cite{kodak} and Tecnick \cite{tecnick} datasets. These approaches generally reduce the redundancy of the images by using an autoencoder architecture, which learns a mapping between the RGB color space and the learned latent space. The latent representation of the image is then quantized into a discrete-valued version, which is further compressed by the lossless entropy coding methods.

Neural data compression methods learn a generalized model to ensure the coding performance during performance evaluation. However, domain shift between the training and testing data and lack of adaptability to the visual content degrade the performance when compressing unseen data samples.
Therefore, some works \cite{content_first,sga,content_lu,content_image} were proposed to improve the adaptability for neural image compression and neural video compression (NVC) by updating the encoder-side components. Those methods aim at generating more compressible latents and estimating more accurate entropy model parameters for each data instance by fine-tuning the latents \cite{content_first,sga}, the encoder \cite{content_lu} or the input image \cite{content_image}. However, such fine-tuning process is extremely time-consuming and the adaptability is still limited due to the fixed decoder.

To exploit the adaptability at the decoder, some full-model over-fitting methods \cite{full} entropy encode and transmit the updates of the decoder parameters along with the quantized latents to the receiver side for better and consistent reconstruction. However, the design of additional model compression is quite complex and the updating approach is also time-consuming. 
Another limitation in NIC is that the number of channels of the latents is not adapted to the rate-distortion (RD) trade-offs and the image content. Most works train multiple models with the same network architecture based on different RD trade-offs for rate control, which generate the latents with the same channel number for different RD trade-offs and spatial locations. However, this leads to redundant elements in the latents.

In this work, we propose a content adaptive NIC framework to improve the adaptability on both latents and the decoder. To improve the adaptability of latent codes, we propose the content adaptive channel dropping (CACD) method, which selects the optimal quality level at each spatial location for the latents and drops redundant elements along the channel dimension. In order to improve decoder-side content adaptability, we propose the content adaptive feature transformation (CAFT) method for the decoder, which extracts characteristic information of the image content in the decoder side and utilizes it to adapt each upsampled feature to the image content by using the Spatial Feature Transform (SFT) \cite{sft_isr} strategy.

The experiments demonstrate that our proposed methods improve the performance of the baseline framework \cite{nips18} in terms of both latents and the decoder. Our proposed content adaptive methods are also complementary to those encoder-side updating methods. Experimental results on the Kodak dataset demonstrate that our framework equipped with the encoder-side updating method Stochastic Gumbel Annealing (SGA) \cite{sga} achieves comparable overall results to the recent state-of-the-art NIC methods \cite{acm21,lbhic} and outperforms them in terms of PSNR.
Additionally, the experimental results also indicate that our methods are general and can be readily applied to NVC for better coding performance. The contributions of our work are summarized as follows:

\begin{itemize}
\item We propose the content adaptive channel dropping (CACD) method to improve the adaptability of RD trade-offs and the image content for latent codes. Our CACD automatically selects the optimal quality level at each spatial location, and then drops redundant elements for bit-rate saving.
\item To exploit the adaptability at the decoder side, our content adaptive feature transformation (CAFT) method modulates the output features at multiple levels by considering the characteristic information of the image content.
\item Experimental results demonstrate that our methods improve the performance by adapting both latents and the decoder without any additional updating steps during performance evaluation, which are also complementary to the encoder-side updating methods.
\end{itemize}

\section{Related Work}
\subsection{Neural Image Compression}
In recent years, neural image compression (NIC) performance has been improved significantly, which are mostly based on recurrent neural networks (RNNs) \cite{rnn1,rnn2,rnn3}, convolutional neural networks (CNNs) \cite{iclr17,iclr18,nips18,cvpr20,lbhic,chen_img,iclr22}, or invertible neural networks (INNs) \cite{acm21}. 
In most works, CNN-based autoencoder is selected as the basic framework. Ball\'{e} \textit{et al.} \cite{iclr17} proposed an end-to-end optimized image compression framework based on nonlinear transformation, the additive noise quantization proxy and the fully factorized entropy model. 
Subsequently, the researchers focus more on improving the accuracy of the estimated entropy model using hyperprior \cite{iclr18}, auto-regressive context model \cite{nips18} and Gaussian Mixture Model (GMM) \cite{cvpr20}. 
Different transformations are also proposed to enhance the expression capability of the latent space, such as residual blocks with attention module \cite{cvpr20} and INN \cite{acm21}.
Some works \cite{rnn1,importance} applied the spatially variant bit allocation strategy as a post-process \cite{rnn1} or by using importance map \cite{importance}.
Our method is also based on the convolutional autoencoder approach, but we improve the content adaptability of the baseline method \cite{nips18}.

\subsection{Content Adaptive Data Compression}
The effectiveness of neural data compression relies on the generalization capability to unseen data in the evaluation process.
However, domain shift between training and testing data and lack of adaptability may degrade the coding performance when compressing various types of testing data. To solve this issue, a straightforward idea is to over-fit the encoder-side components. In this way, the model can adapt to test samples during performance evaluation, and does not affect the reconstruction quality because the encoder is not involved in the decoding process. 
To this end, Campos \textit{et al.} \cite{content_first} refined the latents by directly back propagating them, 
and Yang \textit{et al.} \cite{sga} further closed the discretization gap by replacing the differentiable approximation for quantization with Stochastic Gumbel Annealing (SGA) when refining the latents. 
Moreover, Lu \textit{et al.} \cite{content_lu} updated the encoder on each test frame for neural video compression (NVC), which generates content adaptive latent codes by using the over-fitted encoder.
Recently, some full-model adaption methods for NVC have been proposed to adapt the decoder. The work in \cite{full} updated both encoder and decoder when compressing I frames, and then transmitted the updates of the decoder parameters along with the compressed video sequences. 
These updating methods require hundreds or thousands of back propagation steps for each sample, which is extremely time-consuming. In summary, the encoder-side approaches do not utilize the adaptability of decoders and the full-model approach is often complex due to the additional model compression process.

Our proposed content adaptive methods adapt the latents and the decoder to the image content in a non-updating way. Our methods are also complementary to those encoder-side updating methods, which leads to a fully-adapted solution to address the issues of both domain shift and lack of adaptability.

\subsection{Neural Video Compression}
In recent years, significant progress has also been achieved for neural video compression (NVC). Increasing number of learning based approaches \cite{nvc_interpolation,dvc,scale_space_flow,rafc,mlvc,fnvc_1,fnvc_2,fvc,hu2022fvc,chen_vid,hu2022coarse} have been proposed. Lu \textit{et al.} \cite{dvc} first proposed an end-to-end video compression framework DVC that follows the traditional hybrid coding framework and implements the key components with neural networks. Some subsequent works improved the motion compensation \cite{scale_space_flow} or motion compression \cite{rafc} for better optical flow based motion compensation. 
Recently, more works \cite{fnvc_1,fnvc_2,fvc} were proposed to perform the operations in the feature space. Hu \textit{et al.} \cite{fvc} proposed the FVC framework where motion compensation and residual coding are performed in the feature space rather than the pixel space.

\begin{figure}[ht!]
    \centering
    \includegraphics[width=\linewidth]
    {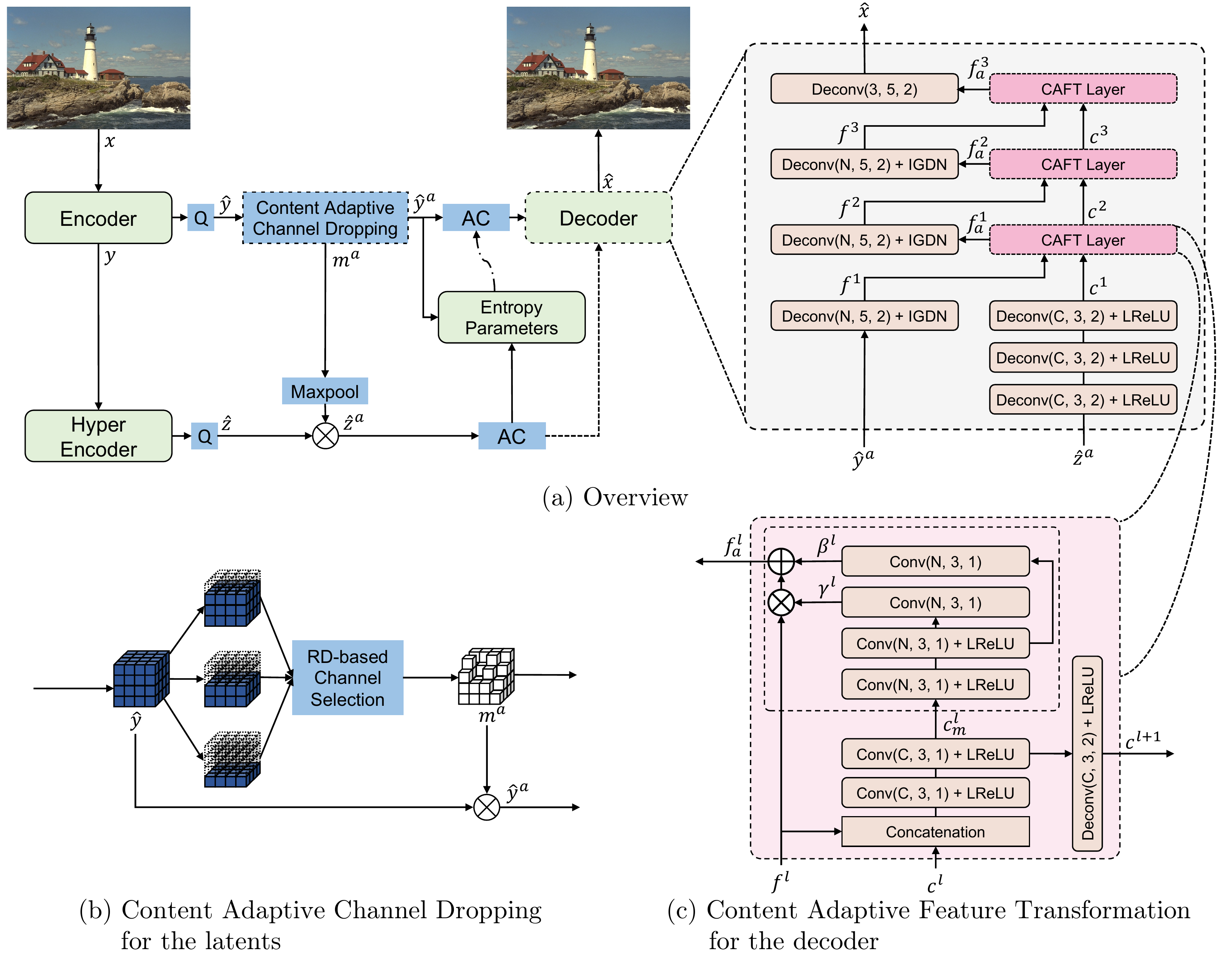}
    \caption{
    Overview of our proposed framework based on \cite{nips18} (a), the details in our content adaptive channel dropping (CACD) method for the latents (b) and the network architecture of our content adaptive feature transformation (CAFT) method for the decoder (c). For simplicity, the hyper-decoder and auto-regressive context model are denoted as ``Entropy Parameters" and AC denotes arithmetic coding in the pipeline (a). The operation and modules with dashed container (\textit{i.e.}, the CACD, the CAFT and the decoder in (a)) along with the dashed data flow are our newly proposed modules. In CACD (b), the features with different channel widths are first generated from the quantized latent representation $\hat y$. Then the rate-distortion (RD) based selection technique is applied to select the optimal channel number for each spatial location, which is stored in a binary mask $m^a$. Channel dropping is then completed by element-wise multiplication of the latents $\hat y$ and the mask $m^a$, which is also used to generate $\hat z^a$ (see section~\ref{sect:impo} for more details). 
    In CAFT (c), we adapt each upsampled feature with the  transformation parameters generated by using the Spatial Feature Transform (SFT) layer conditioned on the characteristic information of the image content. The characteristic information is first generated by using the hyperprior $\hat{z}^a$ in the decoder and then mixed with intermediate features and upsampled in CAFT layers (see section \ref{sect:decoder} for more details). Conv(C, K, S) denotes the convolution layer with the output channel $C$, the kernel size $K\times K$ and the stride $S$. LReLU denotes the LeakyReLU activation for simplicity.
    }
    \label{fig:architecture_overall}
\end{figure}

\section{Proposed Method}
\subsection{Overall Architecture of Neural Image Compression}
\label{sect:overall}
We use the state-of-the-art neural image compression (NIC) method \cite{nips18} as our baseline method and apply our methods on top of both context version and the non-context version. The overview of the baseline framework is provided in Fig.~\ref{fig:architecture_overall}(a). We also describe the details of the baseline method as follows.

At the encoder side, the input image $x$ is first transformed into the latent representation $y$ by using the encoder network, which consists of several convolution layers and uses the generalized divisive normalization (GDN) \cite{gdn} layer as activation. The hyper-encoder captures the spatial dependencies of $y$ and produces the hyperprior $z$. 
Then $y$ and $z$ are quantized into discrete-valued version $\hat y$ and $\hat{z}$ respectively
by using the round operation, which is replaced by adding uniform noise \cite{iclr17} as an approximation during the training process. After that, the quantized features $\hat{y}$ and $\hat{z}$ are entropy coded into bit-stream.
Each element in $\hat z$ is modeled as a factorized model $p_{\hat{z}}$ and each element in $\hat y$ is modeled as a Gaussian distribution $p_{\hat{y} \mid \hat{z}}$ conditioned on $\hat{z}$.

At the decoder side, the quantized hyperprior $\hat z$ is first entropy decoded and used to estimate the distribution of the quantized latent representation $\hat y$. In the non-context version of \cite{nips18}, $\hat z$ is fed into the hyper-decoder to estimate the mean and standard deviation of $\hat y$. While in the context version, an auto-regressive context model is added  to utilize the entropy-decoded parts of $\hat y$ for more accurate entropy parameter estimation. Finally, the decoder takes  $\hat y$ as the input to generate the reconstructed image $\hat x$ by using several deconvolution layers and inverse generalized divisive normalization (IGDN) layers.

During the training process of NIC, a rate-distortion optimization (RDO) problem is formulated to minimize the bit-rate cost and the distortion between the original image $x$ and its reconstruction image $\hat x$. 
A Lagrange multiplier $\lambda$ is used to control the trade-off between the bit-rate cost and the distortion. The loss function is formulated as follows:
\begin{align}
    R+\lambda D & = H(\hat y) + H(\hat z) + \lambda d(x, \hat x)
    \label{eq:rdo}
\end{align}
where $H(\hat y)$ and $H(\hat z)$ denote the bit costs to compress $\hat y$ and $\hat z$, $d(x, \hat x)$ denotes the distortion between the reconstructed image and the input image, where mean squared error (MSE) is usually used.

In our approach, we propose new operations and modules for the latents and the decoder. The channel dropping algorithm selects the optimal quality level at each spatial location for the latents $\hat y$ by minimizing the rate-distortion (RD) value. Then the latents $\hat y$ and the hyperprior $\hat z$ are replaced with their channel-adapted version $\hat y^a$ and $\hat z^a$ before entropy coding for bit-rate saving, where the exceeding channels are dropped (see section \ref{sect:impo} for more details).
In the decoder, we modulate the upsampled features after each IGDN layer by using the Spatial Feature Transform (SFT) \cite{sft_isr} layer conditioned on characteristic information of the image content, which is first extracted from the hyperprior $\hat{z}^a$ (see section \ref{sect:decoder} for more details).

\subsection{Content Adaptive Channel Dropping for the Latents}
\label{sect:impo}
In neural image compression, rate control is implemented by training the models with different trade-offs (\textit{i.e.}, different $\lambda$ values) between bit-rate cost and reconstruction distortion. It is well-known that the more bits we use, the better reconstruction quality we can achieve. We also observe that the ability of converting extra bits to reconstruction quality (\textit{i.e.}, the quality gain when assigning similar additional bits) is different among image blocks. To this end, we quantify this ability as ``bit conversion ratio", which is formulated as follows:
\begin{align}
    \eta(x, \lambda^{l}, \lambda^{h})_{i} = {{PSNR(x,\lambda^{h})_{i}-PSNR(x,\lambda^{l})_{i}}\over {{R(x,\lambda^{h})_{i}-R(x,\lambda^{l})_i}}}
\end{align}

where $PSNR(x, \lambda)$ denotes the peak signal-to-noise ratio (PSNR) between the input image $x$ and its reconstructed image produced by the model trained with $\lambda$, $R(x, \lambda)$ denotes the bit-rate cost of the latents and the hyperprior generated by the model trained with $\lambda$, $\lambda ^l$ and $\lambda ^h$ denote the relatively lower and higher $\lambda$ values respectively, and $i$ denotes the $i$th spatial block of the image.

\begin{figure}[htbp]
\centering
\includegraphics[width=1\linewidth]{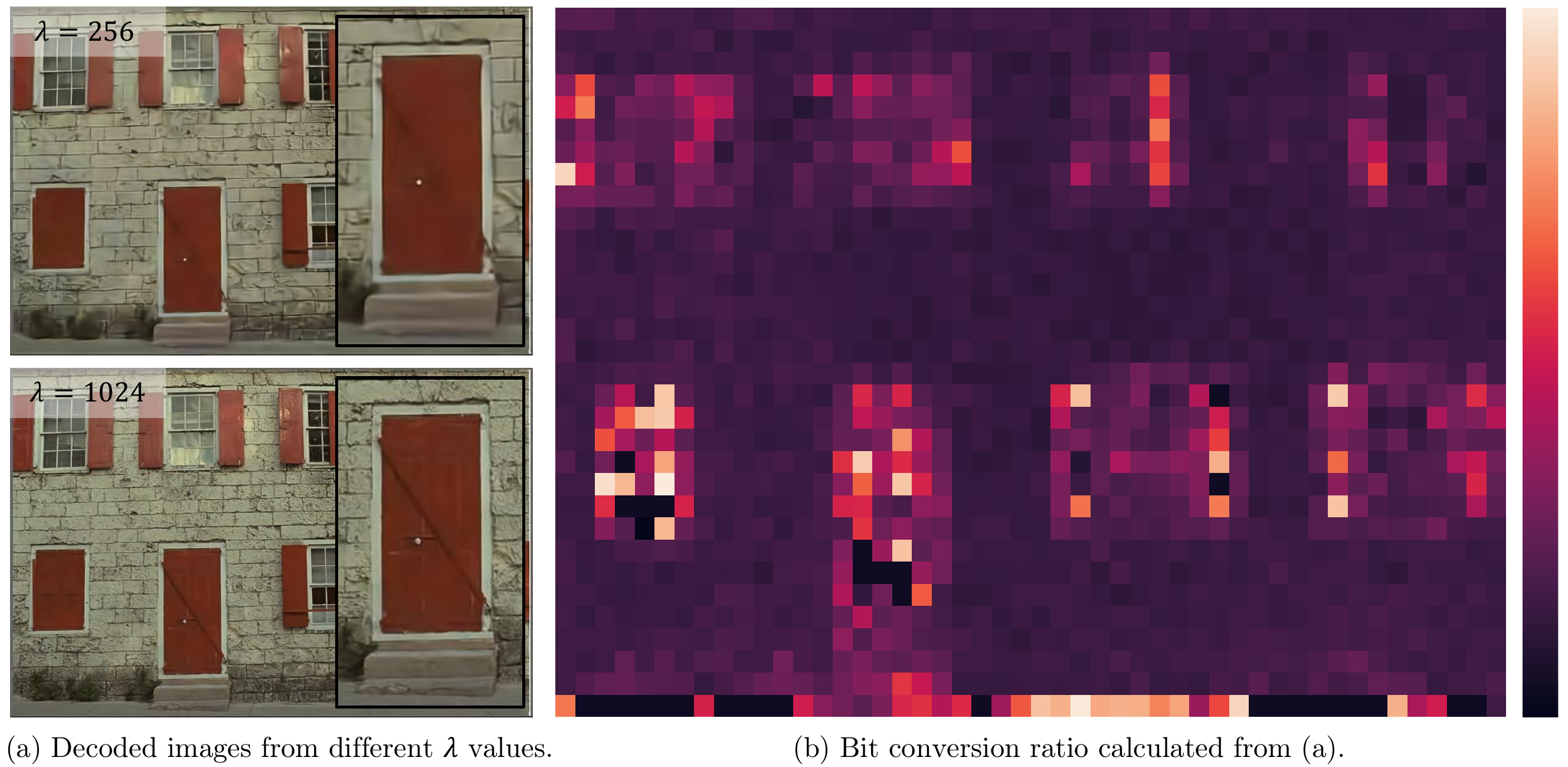}
\caption{An example of bit conversion ratio calculation on an image from the Kodak dataset based on the existing method \cite{nips18}.}
\label{bit_convert_ratio}
\vspace{-0.5cm}
\end{figure}

In Fig.~\ref{bit_convert_ratio}, we provide a visualization example about bit conversion ratio on an image from the Kodak dataset \cite{kodak}. Fig.~\ref{bit_convert_ratio}(a) visualizes two images decoded by  \cite{nips18} trained with two different $\lambda$ values. 
It is observed that the grains of both woodworks  (\textit{i.e.}, the wooden door and windows) and the bricks are constructed with more details in the bottom image with high bit-rate.
In Fig.~\ref{bit_convert_ratio}(b), we observe that the bit conversion ratio of the wooden areas is much higher than that of the brick areas. 
To achieve better rate-distortion (RD) performance, it is therefore reasonable to assign more bits for the areas with higher bit conversion ratio. To this end, we aim at compressing each image block with a suitable quality level, at which the RD cost is minimal among all the alternative quality levels.

Before selecting the quality level at each spatial location for the latents $\hat y$, our content adaptive channel dropping (CACD) method needs to enable multiple quality levels in one single model. For each $\lambda$ value, we first decide the corresponding maximum channel number $g(\lambda)$ (also called the optimal channel width in this work), where the RD performance saturates at this channel width even if more channels are allowed for this $\lambda$ value~\cite{slimmable}. Then we train our model with multiple rate-distortion optimization (MRDO) loss~\cite{slimmable}. Note that we only set the additional elements in the channel dimension as zero instead of directly reducing the number of channels as in the slimmable implementation~\cite{slimmable}. Speciﬁcally, for the original target $\lambda$ value, we have $K$ $\lambda$ values (\textit{i.e.}, $K$ quality levels) including its original $\lambda$ value and $K-1$ smaller $\lambda$ values ($K$ is set as 3 in this work).
A mask $m^{g(\lambda)}$ is generated by setting the value to zero for the channel locations exceeding the channel width $g(\lambda)$, and one otherwise.
The latent representation with level $\lambda$ is generated by the element-wise multiplication operation between the latents $\hat y$ and the corresponding mask $m^{g(\lambda)}$ (\textit{i.e.}, $\hat y^{g({\lambda})} \gets \hat y \odot m^{g(\lambda)}$), and the hyperprior is also mapped in the same way (\textit{i.e.}, $\hat z^{g({\lambda})} \gets \hat z \odot Maxpool(m^{g(\lambda)})$). 
The MRDO loss is then formulated as follows,
\begin{align}
    \sum_{\lambda \in \Lambda} R(\hat y^{g(\lambda)}, \hat z^{g(\lambda)}) + \lambda D(\hat y ^{g(\lambda)})
    \label{mrdo_loss}
\end{align}
where $\Lambda$ denotes the set of $K$ $\lambda$ values, $R$ and $D$ denote the rate cost and the distortion in Eq.(\ref{eq:rdo}) respectively, and they are calculated by using the features with different quality levels. 

As the model can compress the image with $K$ quality levels, we adopt the block-based RD selection strategy, which selects the optimal channel width among alternatives for the smallest RD value at each spatial location. 
Specifically, at each spatial location, we calculate $K$ RD values by using the features with the channel widths $g(\lambda)$ among alternative quality levels and store the channel width corresponding to the smallest RD value in the channel allocation matrix $a$. We further generate the adaptation mask $m^a$ by setting the value to zero for the channel locations exceeding the allocated channel width, and one otherwise.
Then the adapted features are generated by the element-wise multiplication operation with the adaption mask $m^a$ (\textit{i.e.}, $\hat y^a \gets \hat y \odot m^a, \hat z^a \gets \hat z \odot Maxpool(m^a)$).
Therefore, our CACD method for the latents can automatically drop redundant elements at each spatial location and thus reduce the bit-rate cost.

\subsection{Content Adaptive Feature Transformation for the Decoder}
\label{sect:decoder}
Domain shift between the training and testing data is a common problem for learning-based algorithms.
Different from most tasks, the ground truth in neural image compression is exactly the same as the input image. Thus the model can be fine-tuned with the whole target domain dataset or even a target sample. Generally, only the encoder-side components are adapted because the change in the decoder will result in inconsistent reconstruction at the receiver side, which can not exploit the adaptability in the decoder. Although some works \cite{full} synchronize the decoder to the receiver by transmitting the parameter changes, it is a non-trivial task to compress such parameter changes.

Recently, Spatial Feature Transform (SFT) \cite{sft_isr} has shown efficient spatial adaptability for various vision tasks including image super-resolution \cite{sft_isr}, semantic image synthesis \cite{sft_semantic_image_synthesis} and variable-rate image compression \cite{sft_variable_rate}. Inspired by these works, we propose the content adaptive feature transformation (CAFT) method for the decoder, which uses the SFT layers conditioned on the relatively high-level characteristic information to adapt the decoder to the image content.

Fig.~\ref{fig:architecture_overall}(a) shows the architecture of the decoder network with our proposed CAFT layers. In the decoder, we first extract the characteristic information of the image content from the hyperprior $\hat z$ by using the image characteristic extractor, which consists of 3 transposed convolution layers (C is set as 192 in this work).
To adapt the features in the decoder, we append our proposed CAFT layer after each IGDN layer, which considers the characteristic information $c^l$ as the condition of the SFT layer to adapts the up-sampled feature $f^l$ and produces the characteristic information $c^{l+1}$ at next level, where $l=1...L$ (L is set as 3 in this work).

The architecture of the CAFT layer is shown in Fig.~\ref{fig:architecture_overall}(c).
In the CAFT layer at level $l$, the characteristic information $c^{l}$ and the feature to be adapted $f^l$ are first concatenated in the channel dimension and mixed by using several convolution layers. 
The mixed characteristic information $c_m^l$ then forwards two-fold.
On one hand, it is up-sampled to $c^{l+1}$ if $l$ is not the last level. On the other hand, it is input into the conditioned SFT layer, which generates the affine transformation parameters $(\gamma ^l, \beta ^l)$ by learning the mapping function $\Psi(c^l) \mapsto(\gamma^l, \beta^l)$. The input feature $f^l$ is then transformed by using the learned parameters $(\gamma ^l, \beta ^l)$ to produce the content adapted feature $f^l_a$:

\begin{align}
f^l_a = f^l \odot \gamma ^ l + \beta ^ l
\end{align}
where $\odot$ denotes the element-wise multiplication operation.

Our CAFT modulates the features by using the conditioned SFT layer whose condition is the relatively high-level characteristic information of the image content to improve decoder-side content adaptability.

\section{Experiments}
\subsection{Experimental Setup}
\subsubsection{Datasets.}
We adopt the Flicker 2W dataset from \cite{training_set} as our training dataset, which consists of 20,745 images. Each image is randomly cropped into $256\times256$ patches for data augmentation. The rate-distortion performance of our method is evaluated on the Kodak \cite{kodak} and Tecnick \cite{tecnick} datasets.

\subsubsection{Implementation Details.}
We apply our proposed content adaptive methods on both \cite{nips18} and its non-context version. 
We train our models with seven $\lambda$ values (\textit{i.e.}, $\lambda = $ 128, 256, 512, 1024, 2048, 4096 and 6144). We use $N$=$M$=192 for the three lower $\lambda$ values and $N$=$M$=320 for the four higher values. We first train two models with higher $\lambda$ values ($\lambda=$ 1024 for low bit-rates and $\lambda=$ 8192 for high bit-rates). Other models are then fine-tuned from their corresponding pretrained model with their $\lambda$ values.

To train our model with content adaptive channel dropping (CACD), we first use the multi rate-distortion optimization (MRDO) technology (Eq.(\ref{mrdo_loss})) to achieve its original performance with multiple quality levels. Then the CACD module is activated to select the optimal channel width in the subsequent fine-tuning iterations.

We use the Adam \cite{adam} optimizer and set the batch size as 4. The initial learning rate is set as $5e-5$. Each fine-tuning step requires 1,000,000 iterations, which uses the initial learning rate for the first 600,000 iterations and $5e-6$ for the remaining iterations. For MS-SSIM \cite{msssim} based rate-distortion performance evaluation, we further fine-tune our model with the learning rate of $5e-6$ for 500,000 iterations by using MS-SSIM as the distortion loss.

\begin{figure}[t!]
    \centering
    \subfigure{
    \includegraphics[width=0.45\linewidth]{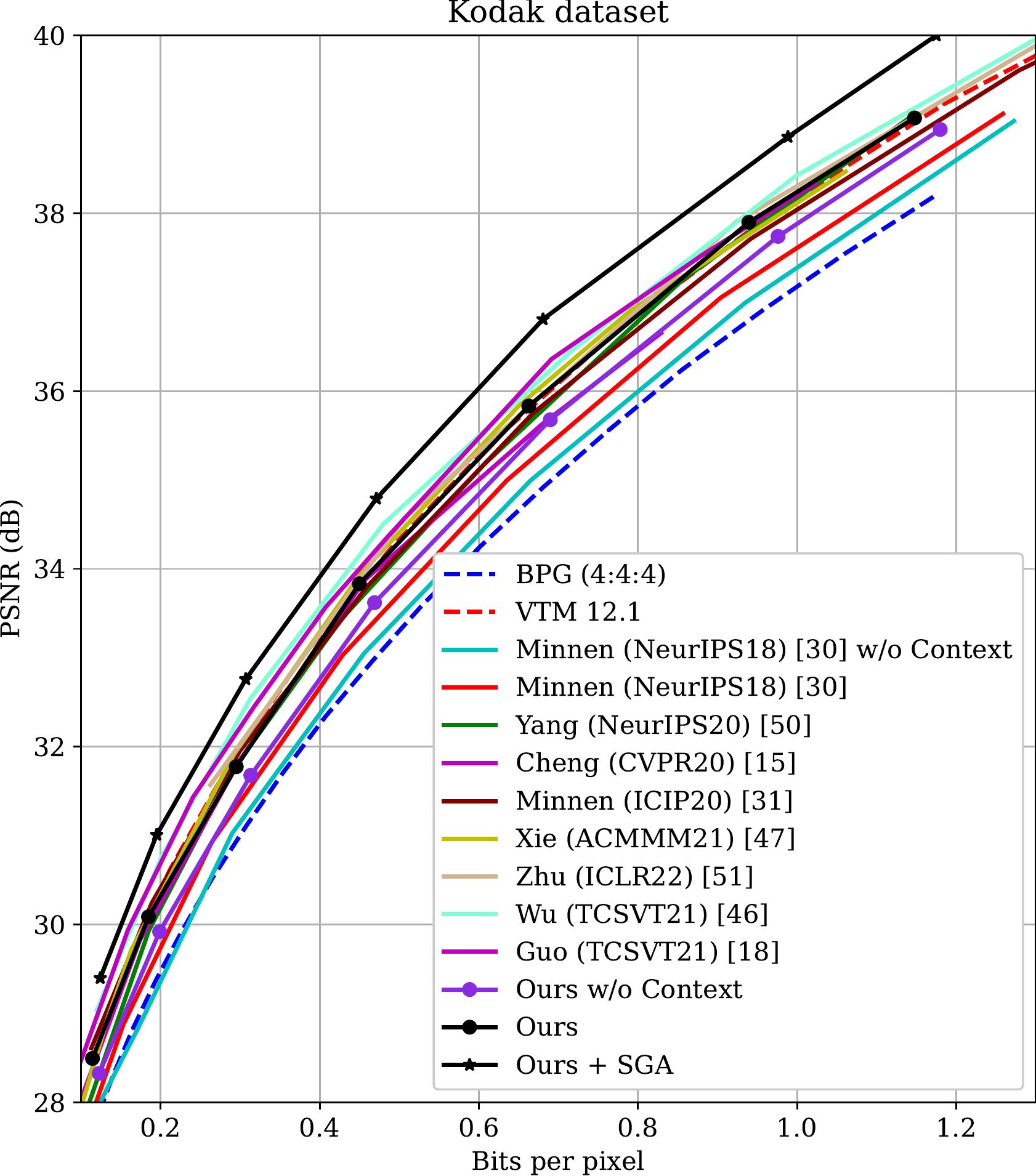}
    }
    \subfigure{
    \includegraphics[width=0.45\linewidth]{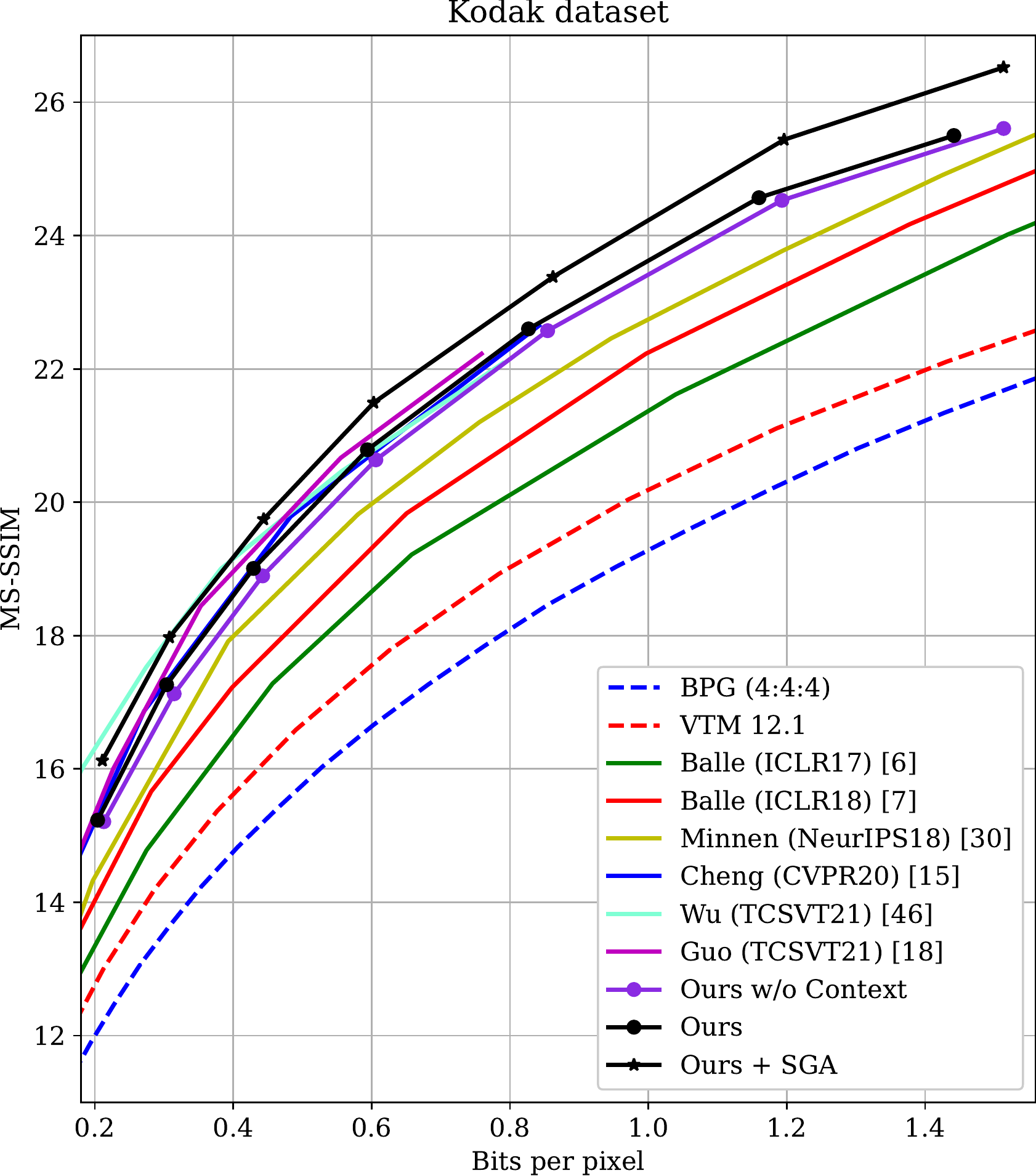}
    }
    \caption{Rate-distortion performance evaluation results on the Kodak dataset.}
    \label{fig:rd_kodak}
\end{figure}

\begin{figure}[t!]
\centering
\begin{minipage}[t]{0.46\textwidth}
\centering
\includegraphics[width=0.99\textwidth]{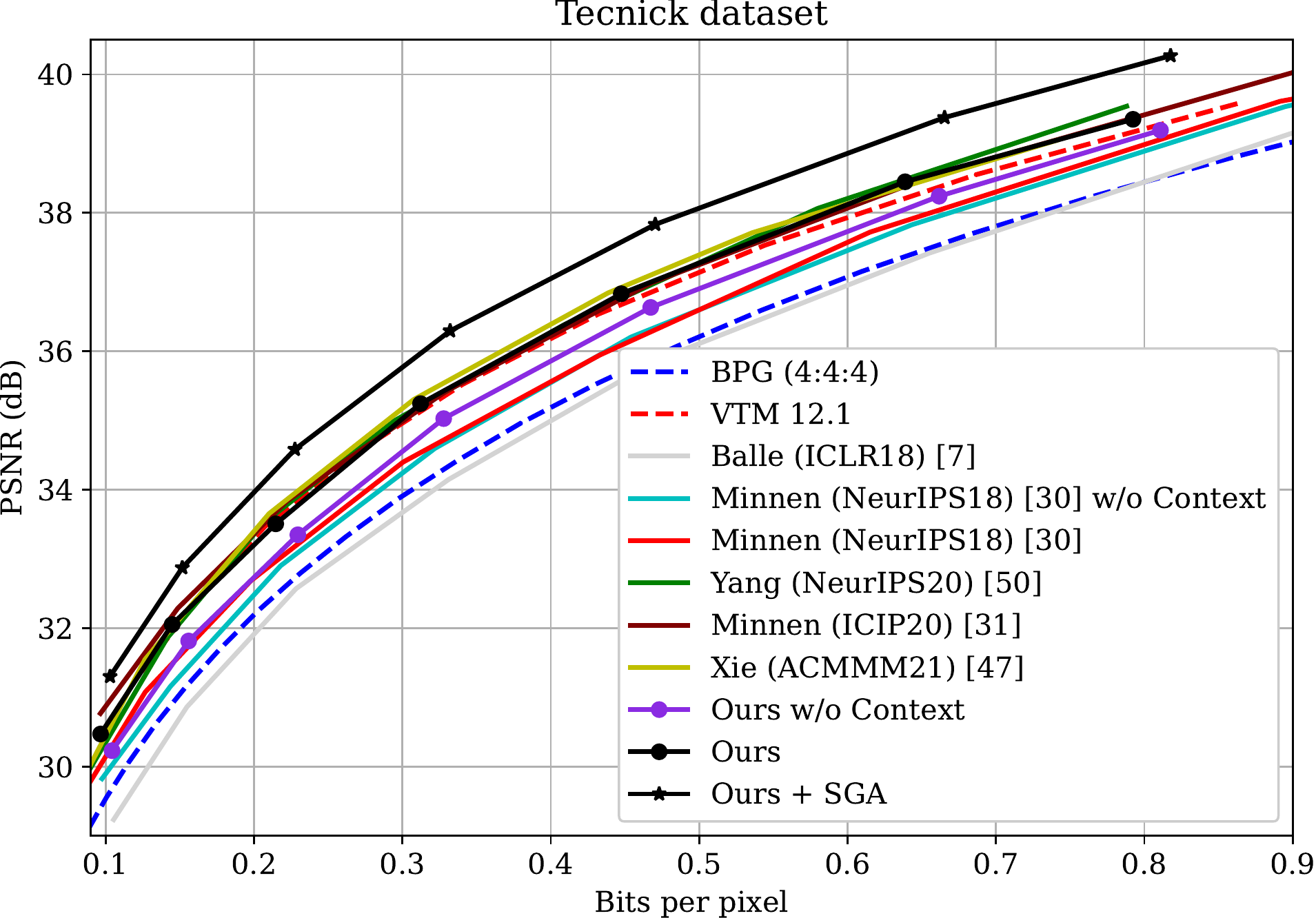}
\caption{Rate-distortion performance evaluation results on the Tecnick dataset.}
\label{fig:rd_tecnick}
\hfill
\end{minipage}
\begin{minipage}[t]{0.46\textwidth}
\centering
\includegraphics[width=0.98\textwidth]{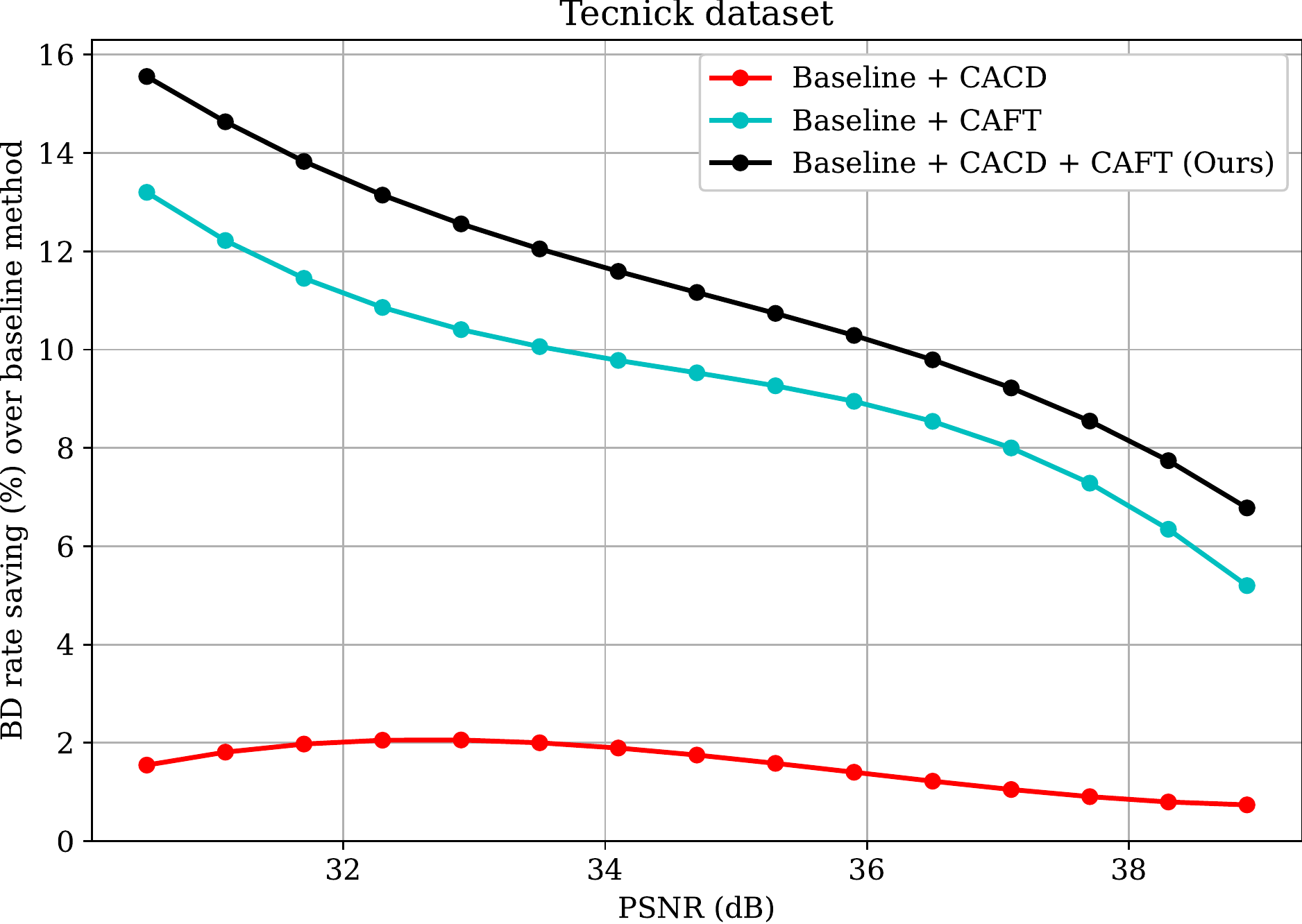}
\caption{BD rate saving (\%) of our CACD and CAFT methods on the Tecnick dataset. We use \cite{nips18} without the auto-regressive context model as our baseline method.}
\label{fig:ablation_tecnick}
\end{minipage}
\end{figure}

\subsection{Rate-Distortion Performance}
In Fig.~\ref{fig:rd_kodak}, we report the performance of traditional image codecs \cite{bpg,vvc}, the state-of-the-art image compression methods \cite{nips18,cvpr20,sga,icip20,acm21,iclr22,tcsvt21,lbhic} and our proposed methods (denoted as ``Ours") on the Kodak dataset. 
VVC is evaluated by VTM-12.1 \cite{vtm} on the CompressAI \cite{compressai}  evaluation platform.
We evaluate our methods on both \cite{nips18} and its non-context version (denoted as a suffix of ``w/o Context"). We observe that our methods improve the rate-distortion performance on both versions of the baseline method in terms of both PSNR and MS-SSIM \cite{msssim}. 
It is worth mentioning that our method is compatible with the state-of-the-art updating-based adaption method Stochastic Gumbel Annealing (SGA) \cite{sga}. We also report the fully-adapted result by combining our methods and SGA, which is denoted as ``Ours + SGA". It is obvious that our fully-adapted method outperforms recent state-of-the-art methods \cite{lbhic,tcsvt21,iclr22} in terms of PSNR. For example, our fully-adapted method achieves 0.4dB improvement at 1.0bpp when compared with the current state-of-the-art methods Xie (ACMMM21) \cite{acm21} and Wu (TCSVT21) \cite{lbhic}.

In Fig.~\ref{fig:rd_tecnick}, we also report the coding performance of different methods on the Tecnick dataset. We have similar observations as on the Kodak dataset that our fully-adapted method achieves the state-of-the-art performance at all bit-rates and achieves 0.6dB improvement at 0.5bpp when compared with current state-of-the-art methods Minnen (ICIP20) \cite{icip20} and Xie (ACMMM21) \cite{acm21}. The experimental results clearly demonstrate the effectiveness of our proposed fully-adapted method.

\subsection{Ablation Study and Model Analysis}
\subsubsection{Effectiveness of the Proposed Methods.}
To demonstrate the effectiveness of our proposed content adaptive methods for the latents and the decoder, we conduct ablation study on the Tecnick dataset. To fairly compare our work with the updating-based adaption method SGA \cite{sga}, we take the non-context version of \cite{nips18} as the baseline method.
We provide the BD rate saving result of our proposed methods over the baseline method based on the piecewise BDBR \cite{bdbr} results. As shown in Fig.~\ref{fig:ablation_tecnick}, the alternative method equipped with our content adaptive feature transformation (\textit{i.e.}, Baseline + CAFT) outperforms the baseline method with the BD rate saving from 5\% to 13\% at all bit-rates. 
Additionally, the alternative method equipped with our content adaptive channel dropping strategy (\textit{i.e.}, Baseline + CACD) generally achieves better performance than the baseline method.
Our method equipped with both CAFT and CACD achieves the best performance and outperforms all other methods, which saves about 14\% bit-rate in low PSNR range.
We also provide the BDBR \cite{bdbr} results compared with the baseline method in Table \ref{table:bdbr_ablation}, which clearly demonstrates the improvement of our proposed methods over the baseline method.
The ablation study results demonstrate that our overall framework is able to adapt to the image content on both latents and the decoder for better compression performance.

\begin{table}[!t]
\begin{center}
\caption{BDBR(\%) results of our proposed methods on the Tecnick dataset. Negative values indicate bit-rate saving. We use \cite{nips18} without the auto-regressive context model as the baseline method to calculate the BDBR results.}
\label{table:bdbr_ablation}
\begin{tabular}{ll}
\hline\noalign{\smallskip}
Methods & BDBR(\%) \\
\noalign{\smallskip}
\hline
\noalign{\smallskip}
Baseline + CACD	& -1.51 \\
Baseline + CAFT	& -9.47 \\
Baseline + CACD + CAFT (Ours) &	-11.08  \\
\hline
\vspace{-1.0cm}
\end{tabular}
\end{center}
\end{table}

\begin{table}[!t]
\begin{center}
\caption{BDBR(\%) results about the compatibility of our method and SGA \cite{sga} on different datasets. Negative values indicate bit-rate saving. We use \cite{nips18} without the auto-regressive context model as the anchor method to calculate the BDBR results.}
\label{table:bdbr_sga}
\begin{tabular}{lll}
\hline\noalign{\smallskip}
Methods & Kodak & Tecnick\\
\noalign{\smallskip}
\hline
\noalign{\smallskip}
SGA & -15.17 & -18.72 \\
Ours w/o Context & -11.51 & -11.08 \\
Ours w/o Context + SGA  & -26.52 & -28.51 \\
\hline
\vspace{-0.6cm}
\end{tabular}
\end{center}
\end{table}

\subsubsection{Compatibility with Updating-based Method in the Encoder Side.}
Our methods adapt to the image content on both latents and the decoder, which is also compatible with the updating-based adaption method SGA \cite{sga}. 
To demonstrate the compatibility, we provide the BDBR \cite{bdbr} results on the Kodak and the Tecnick datasets in Table \ref{table:bdbr_sga}. 
Although our method (\textit{i.e.},``Ours w/o Context") saves less bit-rates than ``SGA", our method in combination with SGA (\textit{i.e.},``Ours w/o Context + SGA") outperforms ``SGA", which indicates that our content adaptive approach is complementary to SGA.

\begin{figure}[t!]
    \centering
    \includegraphics[width=1\linewidth]{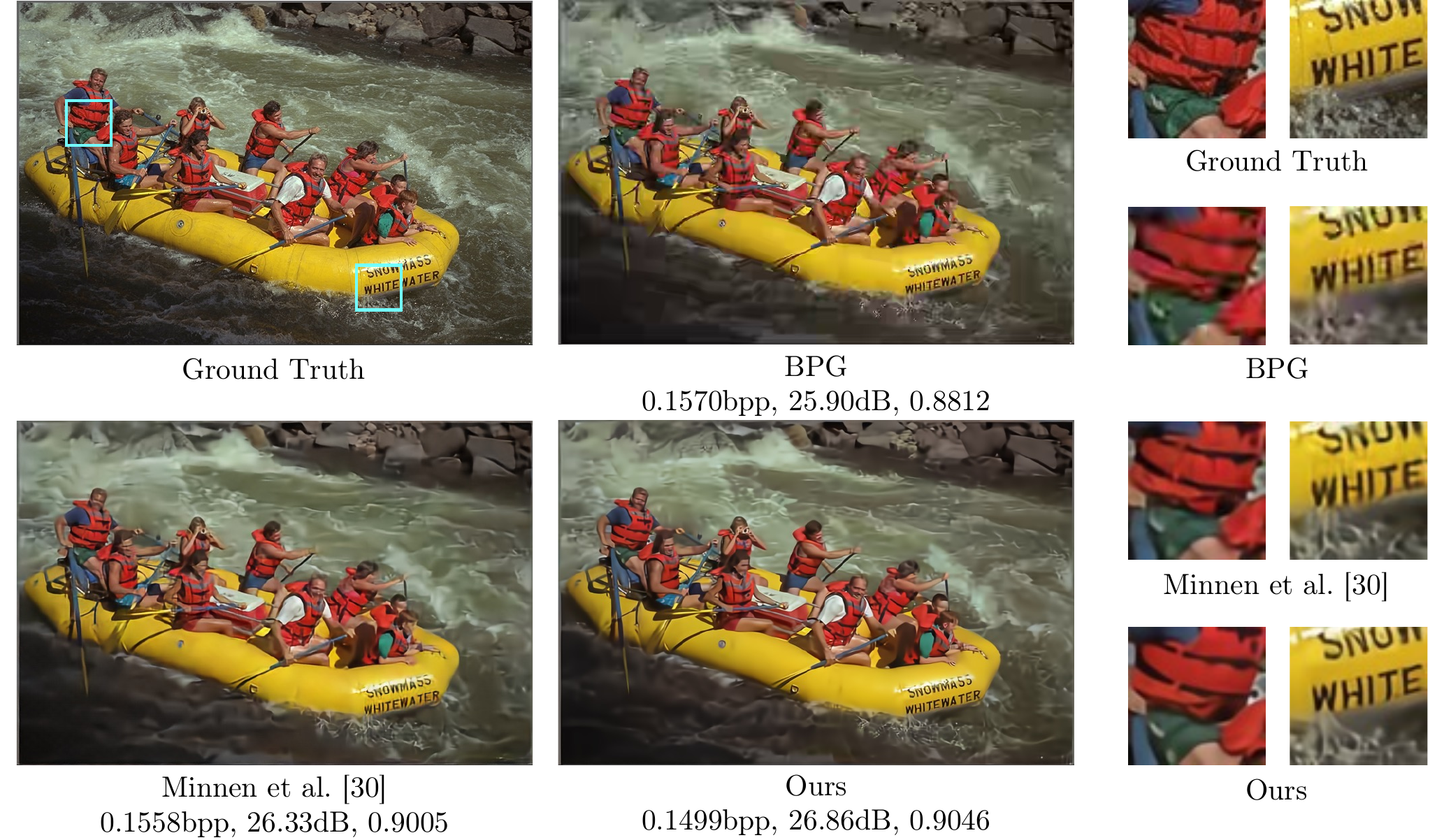}
    \caption{Qualitative comparison results of the traditional codes BPG~\cite{bpg}, neural image compression method Minnen et al.~\cite{nips18} and our method.}
    \label{fig:qualitive_visualization}
    \vspace{-0.5cm}
\end{figure}

\subsubsection{Qualitative Results.}
As shown in Fig.~\ref{fig:qualitive_visualization}, we provide the visualization results of the reconstructed image \textit{kodim14} from the Kodak dataset for qualitative comparison.
It is observed that our method clearly improves the reconstruction quality over the baseline method \cite{nips18} and achieves better performance than BPG. Our method preserves more details of the image content. For example, the artifacts can be clearly observed in both Minnen et al. \cite{nips18} and BPG on the red life jacket, which are less obvious in our method. Additionally, the letters in front of the boat reconstructed by our proposed method are more clear than those reconstructed by other baseline algorithms with similar bit-rates.

\subsubsection{Visualization of Content Adaptive Channel Dropping.}
\begin{figure}[!t]
\centering
\subfigure[Reconstructed image \textit{kodim21} from  \ \newline the Kodak dataset.]{\includegraphics[width=0.48\textwidth]{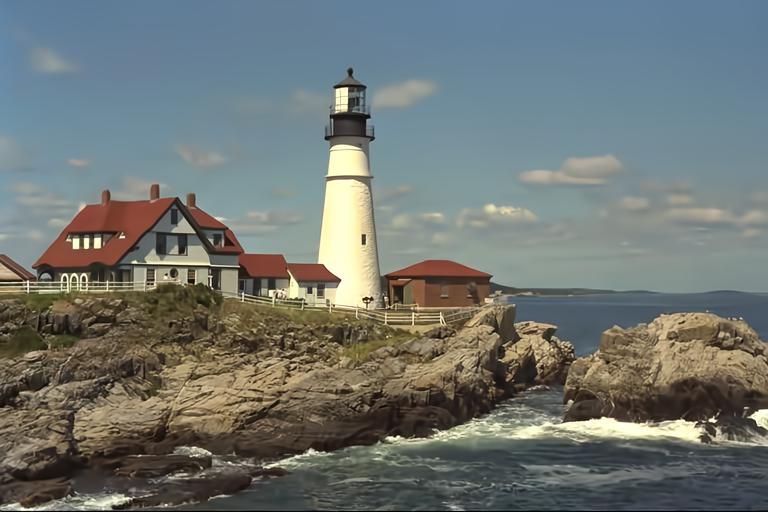}\label{fig:cacd_recon}}%
\subfigure[Visualization of the channel allocation results. Lighter color indicates more channels.]{\includegraphics[width=0.48\textwidth]{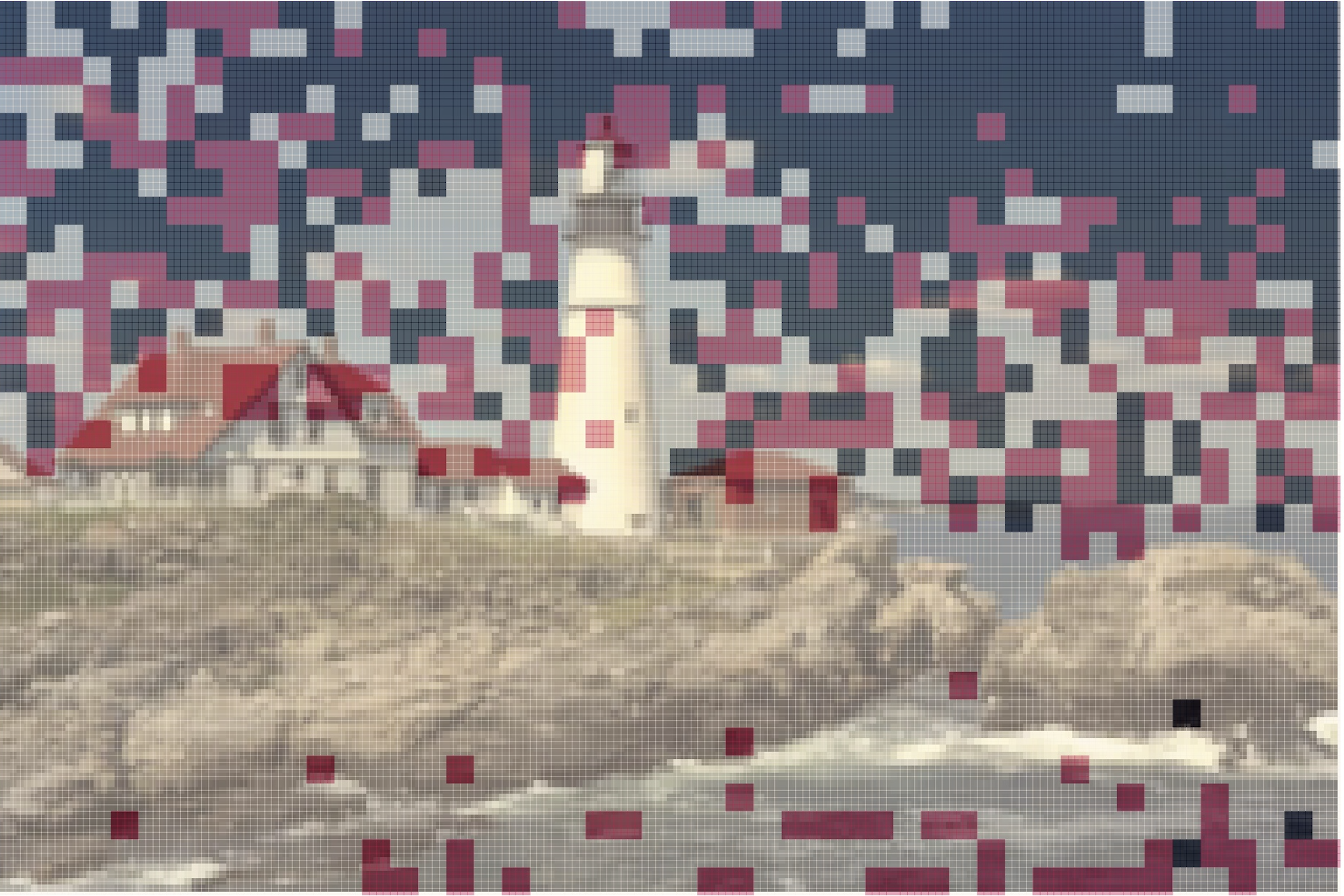}\label{fig:cacd_allocation}}
\caption{Visualization of the channel width selection results by using our method CACD for the latents. In (b), the white, red and black colors represent the quality levels from the highest level (\textit{i.e.}, the most channel number) to the lowest level (\textit{i.e.}, the least channel number).}
\label{fig:cacd_visualization}
\end{figure}

In Fig.~\ref{fig:cacd_visualization}, we visualize the allocated channel number selected by our method CACD. Fig.~\ref{fig:cacd_visualization}(a) is the reconstructed image of \textit{kodim21} from the Kodak dataset and Fig.~\ref{fig:cacd_visualization}(b) visualizes the quality level selection results for the latents. The white, red and black colors represent three quality levels from high to low. It is observed that fewer channels are allocated in the sky area because the sky area is smooth and needs less bits for reconstruction, while full channels are allocated to preserve more details in the sharp areas like the rocks, houses and the lighthouse.

\subsection{Experiments for Neural Video Compression}
\subsubsection{Datasets.}
We train our methods on the Vimeo-90k \cite{vimeo} dataset, which is used as the training dataset in DVC \cite{dvc}. For performance evaluation, we use the video sequences from the HEVC Class B and Class C \cite{h265} datasets.

\subsubsection{Implementation Details.}
We use an enhanced version of DVC \cite{dvc} called ``DVC\textsuperscript{*}" as our baseline method, where the entropy models of both motion vector (MV) feature and residual feature are modeled by the mean-scale hyperprior.
We train the models in a similar way as in neural image compression. We first pretrain a model with the $\lambda$ value of 2048. The learning rate is set as 1e-4 for the first 1,800,000 steps and 1e-5 for the following 200,000 steps. Then we fine-tune the pretrained model with other $\lambda$ values (\textit{i.e.}, 256, 512 and 1024) for 500,000 steps. 
For the adapted model with our proposed content adaptive methods, we fine-tune the baseline model for 400,000 steps with the learning rate as 5e-5 and 100,000 steps with the learning rate as 5e-6 by using different $\lambda$ values.
We use the Adam \cite{adam} optimizer and set the batch size as 4 for all the training procedures.

\begin{figure}[t!]
    \centering
    \subfigure{
    \includegraphics[width=0.47\linewidth]{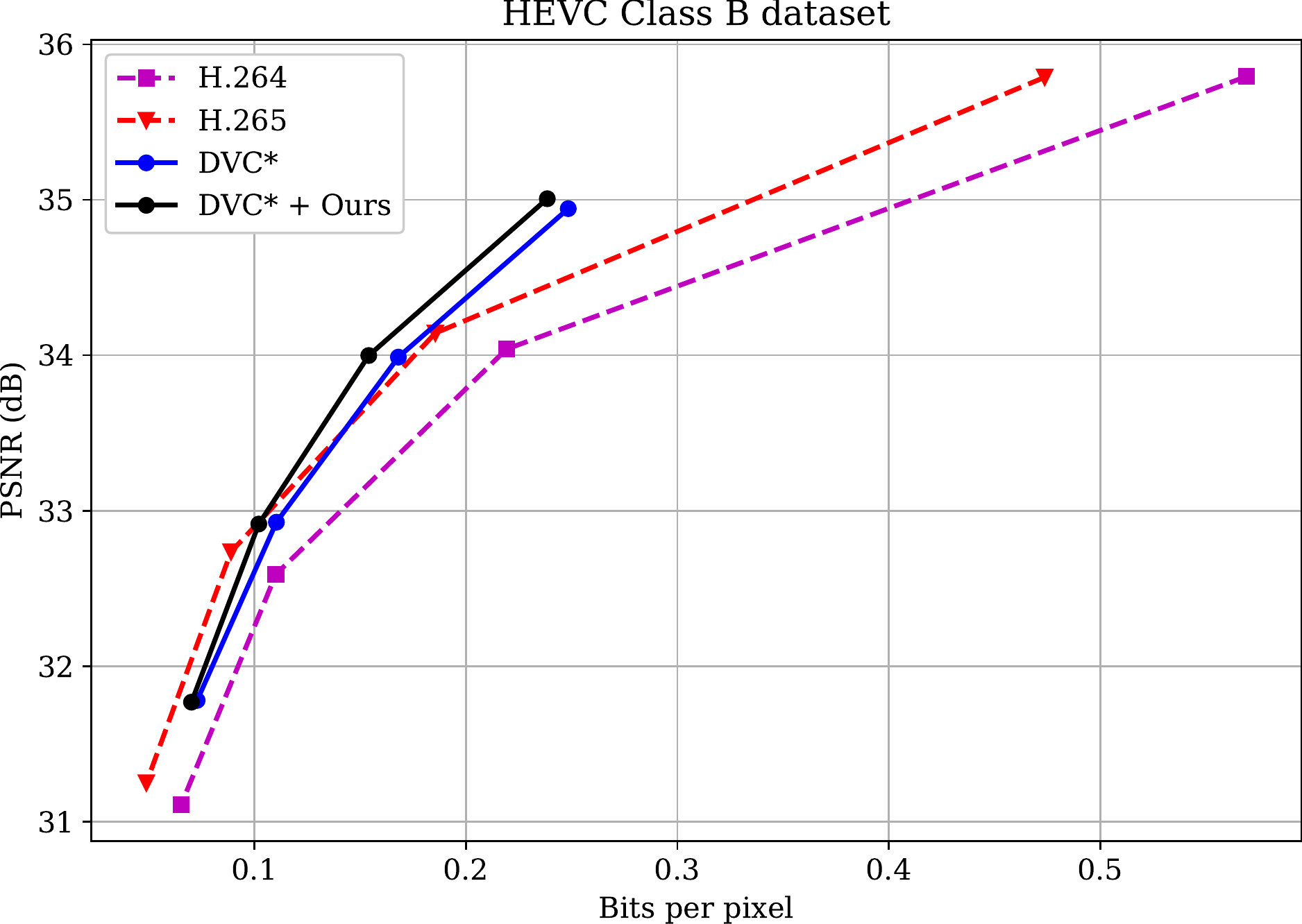}
    }
    \subfigure{
    \includegraphics[width=0.47\linewidth]{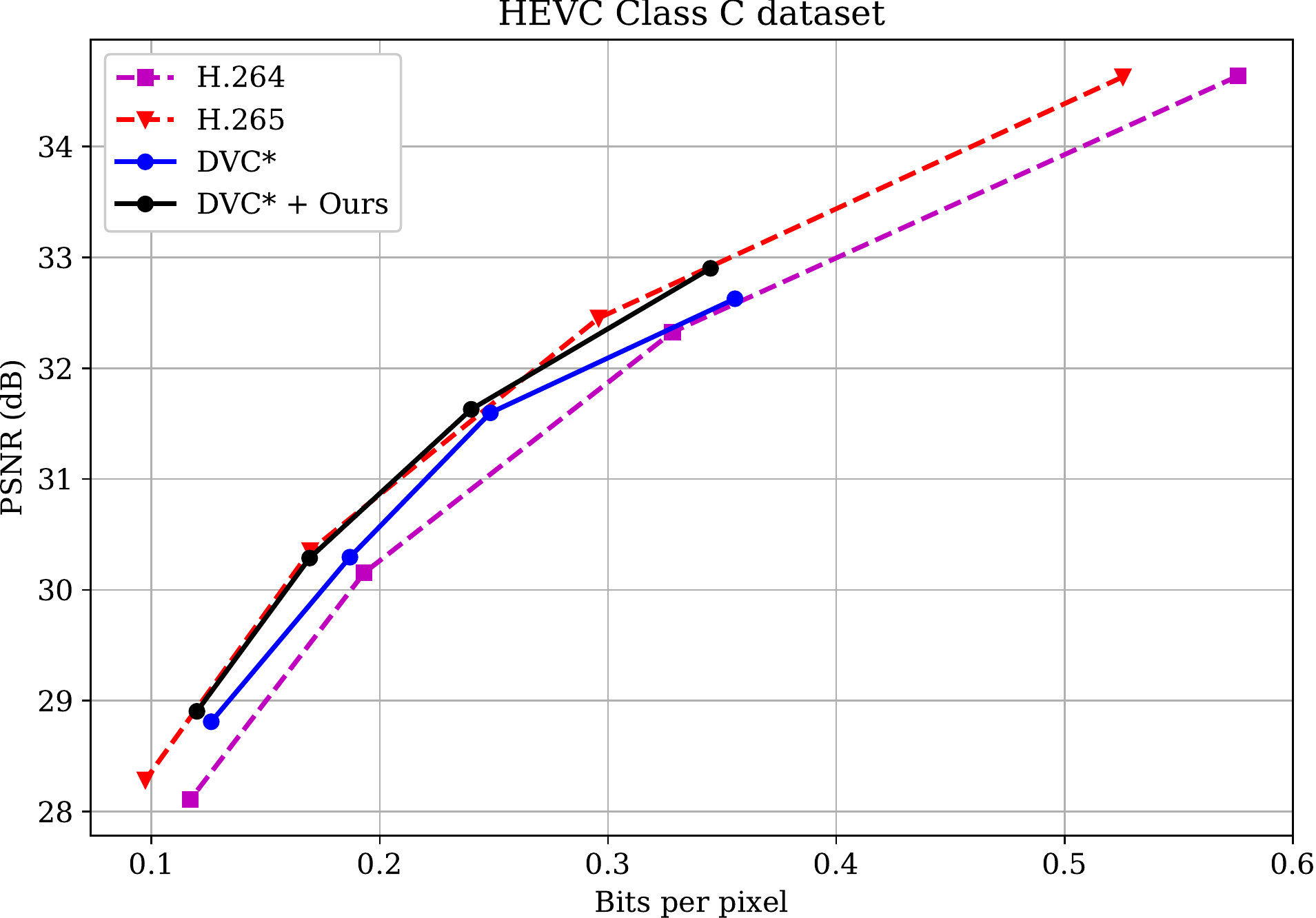}
    }
    \caption{Rate-distortion performance evaluation results on the HEVC Class B and Class C test sequences.}
    \label{fig:rd_video}
    \vspace{-0.5cm}
\end{figure}

\subsubsection{Rate-Distortion Performance.}
Fig.~\ref{fig:rd_video} compares the rate-distortion performance between our methods and the baseline method DVC\textsuperscript{*}. It is observed that our method improves the PSNR by about 0.1 dB at the middle bit-rate and by about 0.3 dB at other bit-rates on the HEVC Class C test sequence. Improvement can also be observed on the HEVC Class B test sequence, which has larger resolution than the HEVC Class C test sequence. The experimental results demonstrate that our methods are general and can be readily used for neural video compression.

\section{Conclusions}
In this work, we have proposed the content adaptive methods for both latents and the decoder to improve the content adaptability for neural image compression. Our newly proposed content adaptive channel dropping (CACD) method is able to adaptively compress different locations with different quality levels by dropping redundant channels for better bit-rate saving. Our newly proposed content adaptive feature transformation (CAFT) method at the decoder side can extract the characteristic information of the image content, which can be further regarded as the condition to transform the features in the decoder. Experimental results demonstrate that our content adaptive methods are general to different compression pipelines and are also complementary to the encoder-side updating-based content adaptive methods.

\section*{Acknowledgement}
This work was supported by the National Key Research and Development Project of China (No. 2018AAA0101900).
\newpage


%
%
\bibliographystyle{splncs04}
\bibliography{egbib}
\end{document}